\documentclass{article}[12]

\usepackage{PRIMEarxiv}
\usepackage[utf8]{inputenc} 
\usepackage[T1]{fontenc}    
\usepackage{hyperref}       
\usepackage{url}            
\usepackage{booktabs}       
\usepackage{amsfonts}       
\usepackage{nicefrac}       
\usepackage{microtype}      
\usepackage{lipsum}
\usepackage{fancyhdr}       
\usepackage{graphicx}       
\graphicspath{{media/}}     

\title{A survey of Identification and mitigation of Machine Learning algorithmic biases in Image Analysis}

\author{
Laurent Risser$^{1,2}$, Agustin Picard$^{2,3}$, Lucas Hervier$^{4}$, Jean-Michel Loubes$^{1,2}$\\
$\null$\\
$\null^{1}$ Institut de Math\'ematiques de Toulouse (UMR 5219), CNRS, Université de Toulouse, F-31062 Toulouse, France\\
  $\null^{2}$ Artificial and Natural Intelligence Toulouse Institute (ANITI), Toulouse, France\\
  $\null^{3}$ Scalian, Lab\`ege, France\\
  $\null^{4}$ Institut de Recherche Technologique (IRT) Saint Exup\'ery, Toulouse, France\\
}

\begin{document}
\maketitle

\begin{abstract}
The problem of algorithmic bias in machine learning has gained a lot of attention in recent years due to its concrete and potentially hazardous implications in society.
In much the same manner, biases can also alter modern industrial and safety-critical applications where machine learning are based on high dimensional inputs such as images. This issue has however been mostly left out of the spotlight in the machine learning literature. Contrarily to societal applications where a set of proxy variables can be provided by the common sense or by regulations to draw the attention on potential risks, industrial and safety-critical applications are  most of the times sailing blind. The variables related to undesired biases can indeed be indirectly represented in the input data, or can be unknown, thus making them harder to tackle. This raises serious and well-founded concerns towards the commercial deployment of AI-based solutions, especially in a context where new regulations clearly address the issues opened by undesired biases in AI. Consequently, we propose here to make an overview of recent advances in this area, firstly by presenting how such biases can demonstrate themselves, then by exploring different ways to bring them to light, and by probing different possibilities to mitigate them. We finally  present a practical remote sensing use-case of industrial Fairness.

\end{abstract}


\section{Introduction}\label{sec:intro}

The ubiquity of Machine Learning (ML) models, and more specifically deep neural network (NN) models, in all sorts of applications has become undeniable in recent years. From classifying images~\cite{deng2009imagenet, lecun2010mnist, helber2019eurosat}, detecting objects~\cite{COCO,deng2009imagenet} and performing semantic segmentation~\cite{cordts2016cityscapes,COCO} to translating from one human language to another~\cite{Koehn2005EuroparlAP} and doing sentiment analysis~\cite{imdb-reviews}, the advances in different subfields of ML can be attributed mostly to the explosion of computing power and their ability to speed up the training process of artificial NNs. Most famously, AlexNet~\cite{alexnet} allowed for an impressive jump in performance in the challenging ILSVRC2012 image classification dataset~\cite{deng2009imagenet}, also known as ImageNet, permanently cementing deep convolutional NN (CNN) architectures in the field of computer vision. Since then, architectures have gotten more refined~\cite{vgg, resnet}, training procedures have gotten increasingly more complex~\cite{batchnorm}, and their performance and robustness have greatly improved as a consequence.
Namely, the success of these deep CNN models is related to their ability to treat high-dimensional and complex data such as images or natural language.
The impressive performance of NNs for machine learning tasks can be explained by the ability of their flexible architecture to capture meaningful information on various kinds of complex data and the fact that they are potentially composed of millions of parameters. 

\begin{figure}
\includegraphics[width=\linewidth]{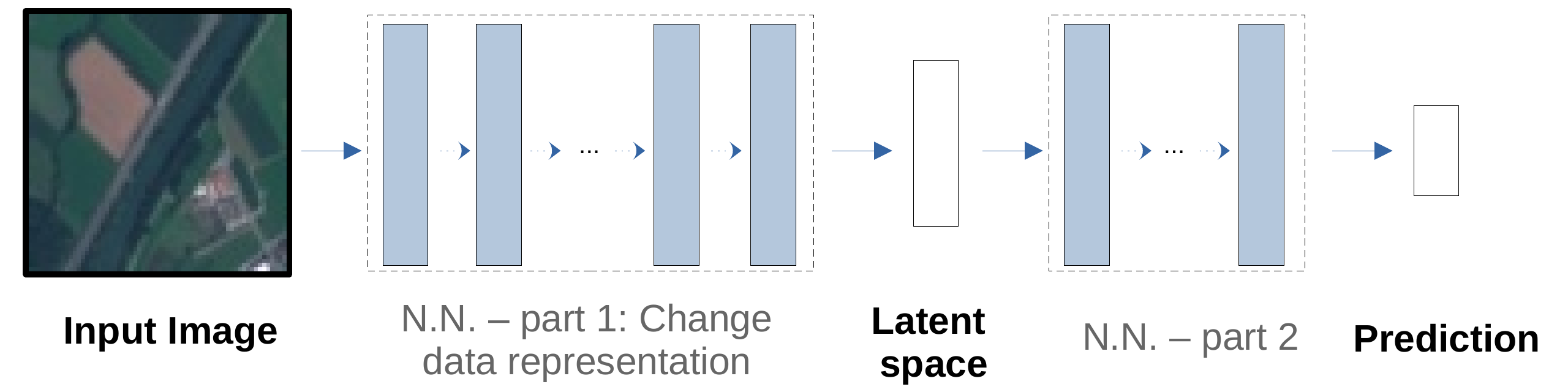}
\caption{General architecture of a neural network designed for classification or regression tasks on images. It first non-linearly projects the input image information into a latent space, and then uses this transformed information for its prediction.}
\label{fig:generalNNarchitecture}
\end{figure}

However, this poses a major challenge: deciphering the reasoning behind the model's predictions. For instance, typical NN architectures for classification or regression problems incrementally transform the representation of the input data in the so-called \textbf{latent  space} (or \textbf{feature space}) and then use this transformed representation to make their predictions, as summarized in Fig.~\ref{fig:generalNNarchitecture}.
Each step of this incremental data processing pipeline (or feature extraction chain) is carried out by a so-called layer, which is mathematically a non-linear function (blue rectangle in Fig.~\ref{fig:generalNNarchitecture}). It is typically made of a linear transformation followed by a non-linear activation function \cite{lecunEtAl1998,vgg}, but more complex alternatives exist -- \textit{e.g.} the residual block layers of ResNet models \cite{resnet} or the self-attention layers \cite{NIPS2017_3f5ee243} in transformer models.
These first stages of the model (Fig.~\ref{fig:generalNNarchitecture}) often rely on the bottlenecking of the information that's passing through it by sequentially decreasing the size of the feature maps and applying non-linear transformations -- e.g. the widely used ReLU activation function \cite{LeCun2012}. To summarize, these first stages project the input data into a latent space.
Therefore, the neural network's data extraction pipeline is driven by the training data that were used to optimize its parameters.
The second part of the network (Fig.~\ref{fig:generalNNarchitecture}), which is standard for classifiers or regressors, is generally simpler to understand than the first, as it is often composed of matrix-vector products (often denoted as dense or fully-connected layers) followed by ReLU activation functions. Consequently, it is mathematically equivalent to a piece-wise linear transformation \cite{arora2018understanding}. More importantly, these non-linear transformations depend on parameters that are optimized to make accurate predictions for a particular task when training the NN.

 Finally, it is worth emphasizing that the data transformation from the latent space to the NN's output can be as complex as in the first part of the network Fig.~\ref{fig:generalNNarchitecture} in models that are not designed for regression or classification, such as \textit{e.g.} the unsupervised auto-encoder models \cite{Kingma2014} or U-Nets \cite{RonnebergerMICCAI2015}. This makes their analysis and control even more complex than in models following the general structure of Fig.~\ref{fig:generalNNarchitecture}. \vskip .1in

The fact that neural networks are \textbf{black-box} models raises serious concerns for applications where algorithmic decisions have life-changing consequences, for instance in societal applications or for high risk industrial systems. This issue has motivated a substantial research effort over the last few years to investigate both explainability, and the creation and propagation of bias in algorithmic  decisions. An important part of this research effort has been made to explain the predictions of black-box ML models  \cite{GradCAM,fel2021sobol,LIME,feature-viz} or to detect out-of-distribution data~\cite{mahalobis-ood,nalisnick2018deep}.\\
\indent In this paper we will leverage the significant work that has been made in the field of \textbf{Fairness}, and study how it can be extrapolated to industrial computer vision applications.
Fairness in Machine Learning considers the relationships between an algorithm and a certain input variable that should not play any role in the model's decision from an ethical, legal or technical point of view, but has a considerable influence on the system's behavior nonetheless. 
This variable is usually called the \textbf{sensitive variable}. Different definitions have been put in the statistical literature, each of them considering specific dependencies between the sensitive variable and the decision algorithm. From a more practical point of view, Fairness issues in Machine Learning manifest themselves in the shape of \textbf{undesired algorithmic biases} in the model's predictions, such as according more bank mortgages to males than females for similar profiles or hiring males rather than females for some specific job profiles, due to a majority presence of male individuals with the corresponding profile in the learning database. Hence, Fairness initially gained a lot of attention specifically in social applications, with a large amount of articles speaking out about the different types of bias that ML algorithms amplify. We refer for instance to the recent review papers of \cite{CastelnovoEtAlNature2022,PessachEtShmueli_ACM_2023} and references therein.

However, we want to emphasize that studies focusing on the presence of bias in more general industrial applications based on complex data like images have mostly been left out of the spotlight. We intend to raise awareness about this kind of problem for safety-critical and/or industrial applications, where trained models may be discriminating against a certain group (or situation) in the form of a biased decision or diminished performance. We point out that a team developing a NN-based application might simply be unaware of this behavior until the application is deployed. In this case, specific groups of end-users may observe that it does not work as intended. A typical example of undesired algorithmic bias in image analysis applications is the one that was made popular by the paper presenting the LIME explainability technique \cite{LIME}. Indeed, the authors trained a neural network to discriminate images representing wolves and huskies. 
Despite the NN's reasonable accuracy, it was still basing itself off spurious correlations -- \textit{i.e.} the presence or not of snow in the background -- to decide whether the image contained a wolf. Another example that will be at the heart of this paper is a blue veil effect in satellite images, which will be discussed in Section~\ref{sec:UseCase}. When present, these biases provide a shortcut for the models to achieve a higher accuracy score both in the training and test datasets, although the logic behind the decision rules is generally false. 
This phenomenon is often modeled by the use of confounding variables in statistics. Hence, they hinder these models' performance when predicting a sample from the discriminated group. 
This makes it completely clear that all harmful biases must be addressed in industrial and safety-critical applications, as algorithmic biases might render the general performance guarantees useless in specific or uncommon situations.

We make the following contribution in this survey: 
\begin{itemize} 
    \item We summarize different types of bias and fairness definitions most commonly present for images.
    \item We present a comprehensive review of methods to detect and mitigate biases with a particular focus on machine learning algorithms devoted to images.
    \item We identify open challenges and discussing future research direction around an industrial use case of image analysis. 
\end{itemize}

\section{Fairness in Machine Learning} \label{s:general}

In this section, we will briefly introduce the different definitions of Fairness that we will consider in this paper. In particular, we will concentrate on statistical -- or global -- notions of Fairness that are the most popular among ML practitioners. There exist other definitions based on causal mechanisms that provide a local measure of discrimination~\cite{KusnerEtAl2017} or \cite{de2021transport} -- and that play an important role in social applications, where discrimination can be assessed individually --, but they are beyond the scope of this paper.

\subsection{Definitions}

Let $X$ be the observed input images, $Y$, the output variables to forecast and $A$, the sensitive variable that induces an undesirable bias in the predictions (introduced in Section \ref{sec:intro}), which can be explicitly known or deduced from $(X,Y)$.
In a supervised  framework, the prediction model $f_{\theta}$ is optimized so that its parameters $\theta$ minimize an empirical risk $R(Y,\hat{Y})$, which measures the error of forecasting $Y$, with $\hat{Y}:= f_{\theta}(X)$. 
We will denote $\mathcal{L}(Z)$ the distribution of a random variable $Z$.

An image is defined as an application $X : K_1  \times K_2  \mapsto \mathbb{R}^d$, where $K_1$ and $K_2$ are two compact sets representing the pixel domain ($K_3$ and $k_4$ can also be considered for 3D or 3D+t images) and $d$ is the number of image channels (\textit{e.g.} $d=3$ for RGB images). We will consider 2D images with $d=1$ in the remainder of this section to keep the notations simple. An image can thus be interpreted as an application mapping each of its coordinates $(i ,j)$ to a pixel intensity value $X(i, j)$.
Metadata, denoted here by ${\rm meta}$, can also be associated to this image. They represent its characteristics or extra information such as the image caption, its location, or even details on the sensor(s) used to acquire or to register it. In a ML setting, the variable to forecast is the output observation $Y$. Fairness is usually assessed with respect to a variable called the sensitive variable $A$ which may be either a discrete variable or a continuous variable. In the discrete case, Fairness objective is to measure dissimilarity in the data and/or discover differences in the algorithm's behavior between samples having different sensitive variable values -- \textit{i.e.} corresponding to different subgroups.
Thus, a complete dataset contains the images $X$, their corresponding target variables $Y$, image metadata ${\rm meta}$ and the sensitive variable $A$. 

Bias can manifest itself in multiple ways depending on how the variable which causes the bias influences the different  distributions of the data and the algorithm.

Bias can originate from the mismatch between the different data distributions in the sense that small subgroups of individuals have different distributions, i.e $\mathcal{L}(Y,X|A) \neq \mathcal{L}(Y,X)$. This is the most common example that we can encounter in image datasets. The first consequence can be a {\bf sampling bias}, and can discourage the model from learning the particularities of the under-represented groups or classes. As a consequence, despite achieving a good average accuracy on the test samples, the prediction algorithm may exhibit poor generalization properties when deployed on real life applications with different subsets of distributions.

Another case emerges when external conditions that are not relevant for the experiment induce a difference in the observed data's labels in the sense that $\mathcal{L}(Y|X,A) \neq \mathcal{L}(Y|X)$, therefore inadvertently encouraging models to learn biased decisions (as in the Wolves versus Huskies example in~\cite{LIME}). This is the case when data is collected with labels influenced by a third unknown variable leading to {\bf confounding bias}, or when the observation setting favors one class over the other leading to {\bf selection bias}. The sources of this bias may be related to observation tools, methods or external factors as it will be pointed out later.

A third interesting case concerns the bias induced by the model itself, which is often referred to as {\bf inductive bias}: $\mathcal{L}(\hat{Y}|X,Y,A) \neq \mathcal{L}(\hat{Y}|X,Y)$. This opposes the {\it world created by the algorithm} --  \textit{i.e.} the distribution of the algorithm outputs -- to the original data. From a different point of view, bias can also arise when the different categories of the algorithm outputs differ from the categories as originally labeled in the dataset -- \textit{i.e.} $\mathcal{L}(Y|\hat{Y},X,A) \neq \mathcal{L}(Y|\hat{Y},X)$ -- a condition that is often referred to as lack of sufficiency. 

Finally, the two previous conditions can also be formulated by considering the distribution of the algorithm prediction errors and their variability with respect to the sensitive variable: $ \mathcal{L}(\ell(Y,\hat{Y})|X,A) \neq \mathcal{L}(\ell(Y,\hat{Y})|X)$, where $ \hat{Y} \times Y \mapsto \ell(\hat{Y},Y)$ is the loss function measuring the error incurred by the algorithm by forecasting $\hat{Y}$ in place of $Y$.

\subsection{Potential causes of bias in Computer Vision}\label{ssec:causes_of_bias}

In practice, the above described situations may materialize through different causes in image datasets.

\subsubsection{Improperly sampled training data}
    First, the bias may come from the data themselves, in the sense that the distribution of the training data is not the ideal distribution that would reflect the desired behavior that we want to learn. Compared with tabular data, image datasets can be difficult to collect, store and manipulate due to their considerable size and the memory storage they require. Hence, many of them have proven to lack diversity -- \textit{e.g.} because not all regions are studied (geographic diversity), or not all sub-population samples are uniformly collected (gender or racial diversity). The growing use of facial recognition algorithms in a wide range of areas affecting our society is currently debated. Indeed, they have demonstrated to be a source of racial~\cite{garvie2016facial,castelvecchi2020facial}, or gender~\cite{conti2022mitigating} discrimination. Besides, well-known datasets such as  CelebA \cite{liu2015faceattributes}, Open Images \cite{KuznetsovaIJCV2020} or ImageNet  \cite{deng2009imagenet} lack of diversity -- as shown in \cite{fabris2022algorithmic} or \cite{shankar2017no} -- resulting in imbalanced samples. Thus, state-of-the-art algorithms are unable to yield uniform performance over all sub-populations.  A similar lack of diversity appear in the newly created Metaverse as pointed out in \cite{riccio2022racial} creating racial bias. This encouraged several researchers to design datasets that do not suffer from these drawbacks -- \textit{i.e.} preserving diversity -- as illustrated by the Pilot Parliament Benchmark (PPB) dataset \cite{buolamwini2018gender} or in \cite{merler2019diversity} or in Fairface dataset \cite{karkkainen2021fairface}.
       
    Combining diverse databases to get a sufficient accuracy in all sub-populations is even more critical for high-stakes systems, like those commonly used in Medicine.
    The fact that medical cohorts and longitudinal databases suffer from biases has been long ago acknowledged in medical studies.
    The situation is even more complex in medical image analysis for specialties such as radiology (National Lung Screening Trial, MIMIC-CXR-JPG \cite{johnson2019mimic}, CheXpert \cite{irvin2019chexpert}) or dermatology (Melanoma detection for skin cancer, HAM10000 database \cite{tschandl2018ham10000}), where biased datasets are provided for medical applications.
    Indeed, under-represented populations in some datasets lead to critical drop of accuracy, for instance in skin cancer detection as in \cite{guo2021bias}, \cite{bevan2021skin} or for general research in medicine \cite{huang2022evaluation} and references therein.
 
    The captioning of images is a relevant example where shortcoming of diversity hampers the quality of the algorithms' predictions, and may result in biased forecasts as pointed out in \cite{wang2022measuring} or in \cite{ross2020measuring}.
    Therefore, it is of utmost importance to include diversity (\textit{e.g} geographic, social, ..) when building image datasets that will be used as reference benchmarks to build and test the efficiency of computer vision algorithms.

\subsubsection{Spurious correlations and external factors}

    The context in which the data is collected can also create spurious correlations between groups of images.
    Different acquisition situations may provide different contextual information that can generate systematic artifacts in specific kinds of images. 
    For instance, confounding variables such as the snowy background in the Wolves versus Huskies example of \cite{LIME} (see Section~\ref{sec:intro}) may add bias in algorithmic decisions. In this case, different objects in images may have similar features due to the presence of a similar context, such as the color background, which can play an important role in the classification task due to spurious correlations. We refer to \cite{singh2020don} for more references. This phenomenon is also well known in biology where spectroscopy data are highly influenced by the fluorescence methods as highlighted in~\cite{0edf374b9737478bb458685cab1b0131}, which makes machine learning difficult to use without correcting the bias. Different biases related to different instruments of measures are also described for medical data in \cite{tschandl2021risk}. \\
    An external factor can also induce biases and shift the distributions. It is important to note that all images are acquired using sensors and pre-processed afterwards, potentially introducing defects to the images. In addition, their storage may require to compress the information they contain in many different ways. All this makes for a type of data with a considerable variability depending on the quality of the sensors, pre-processing pipeline and compression method. 
    This will be illustrated in the application of Section~\ref{sec:UseCase}, where an automatic pre-processing scheme induces a bias in pseudo-color satellite images. In medical image analysis, external factors such as age affect the size of the organs but is also a causal factor to some diseases as analyzed in \cite{pawlowski2020deep}, for instance.

\subsubsection{Unreliable labels}
    We can finally note that wrong or noisy labels, bad captioning (due to stereotyping, for instance) or the use of labeling algorithms that already contain bias (such as Natural Language Processing image interpreters) are also potential source of bias. An example of this phenomenon can be the subjective and socially biased choice of the \textit{attractive} labels in the CelebA \cite{liu2015faceattributes} dataset. When image datasets include captioning as additional variable, the bias inherent to learned language model used to provide the caption is automatically included. For instance one of the main pre-trained algorithm in Natural Language Processing, Generative Pre-Trained Transformer 3, a.k.a GPT-3, is well known to be biased and thus generalized its bias to the image datasets as described for instance in \cite{lucy2021gender} for gender bias.

\subsection{From determined bias to unknown bias in image analysis}\label{ssec:unknown_bias_in_image_analysis}

Keeping in mind the potential sources of bias, different situations may also occur in image analysis applications, depending on the availability of the information:
\begin{itemize}
    \item  \textit{Full information}: images, targets, metadata and sensitive variables, \textit{i.e.} $ (X,Y) \cup \: \{{\rm meta}\} \: \cup A .$ are available.
    The bias may then come from the meta-observations $\{{\rm meta}\}$, the image itself, the labels, or all three.
    \item \textit{Partial information}: the sensitive variable is not observed, so we only observe $(X,Y) \cup \: \{{\rm meta}\}$. The sensitive variable may be included in the meta variables $A \subset \{ {\rm meta} \}$, or may be estimated using the meta-variables $ \{ {\rm meta} \}$.
    \item  \textit{Scarce information}: only the images are observed along with their target, \textit{i.e.} we only observe $ (X,Y)$. The sensitive variable $A$ is therefore hidden. The bias it induces is contained inside the images and has to be inferred from the available data $X$ and used to estimate $A$.
\end{itemize}

For societal applications, the sensitive variable is defined following regulations as presented in Section~\ref{s:droit}. The variable $A$ is known since it is chosen by the regulator, and hence, is either directly available in the data or proxies can be found to estimate it. The main difficulty when working with high dimensional inputs such as images (but also natural language data, time series or graphs) is that the bias may not be explicitly present in a particular input dimension or variable, but is rather hidden in a latent representation of the input data. For instance, an image-based classifier would not naturally have a different rate of positive decisions for males and females because of the intensity of a specific pixel. It would instead detect specific patterns or features in the input images and potentially use this information, leading to unfair decisions with respect to a \emph{gender} sensitive variable. As discussed in Section~\ref{sec:intro}, neural network classifiers or regressors change by construction the representation of the input data into a  lower dimensional latent (or feature) space before making their predictions based on this latent information. Illustrations of how a network can project the input information in a latent space can be found in the VGG and ResNet papers~\cite{vgg, resnet}. It would then be tempting to believe that the hidden variables explaining the undesirable biases would be found in the latent space but this is not necessarily the case. This information can still be distilled in different latent variables unless a specific process is made to isolate it \cite{LocatelloNeurips2019}. Hence, bias detection is an essential, potentially arduous task when dealing with images.

\subsection{Current regulation of AI} \label{s:droit}

It is interesting to remark that the social concerns related to a massive use of AI systems in modern societies has lead to the definition of various ethical and human rights-based declarations  intending to guide the development and the use of these technologies. Some of them were defined by governments or inter-governmental organizations, other ones raised from the civil society, private companies or multi-stakeholders. In 2020, the particularly interesting work of \cite{FjeldEtAl2020} compared the contents of 36 prominent AI principal documents side-by-side. This made clear the similarities and differences in interpretation across these frameworks. This also emphasized the fact that an AI system can be considered as unfair with respect to the ethic principles of one of these documents but fair for another one, which can be particularly confusing for end-users. In order to ensure the trust of the users in AI systems and to properly regulate the use of AI, different states or unions now define specific laws for the use of AI. For instance, the so-called \emph{AI act}\footnote{Proposal for a regulation of the European parliament and of the council laying down harmonised rules on Artificial Intelligence: \url{https://eur-lex.europa.eu/legal-content/EN/TXT/?uri=CELEX\%3A52021PC0206}} of the European Commission will require AI systems sold or developed in the European Union to have proper statistical properties with regard to potential discrimination they could engender (see articles 9.7, 10.2, 10.3 and 71.3). It is also worth mentioning that the article 13.1 of this proposal suggests that the decisions of the AI systems that are likely to pose high risks to fundamental rights and safety (see Annex III of the proposal) may be \emph{sufficiently transparent to enable users to interpret the system’s output}. Making sure that each individual decision can be interpreted by the user is a central question addressed by \textit{explainable AI}, and is the key to understanding whether a specific decision is made by only exploiting pertinent and insensitive information in the input data, or not. As a direct consequence, the user can assess whether an individual decision is fair or not. The sanctions for non-respect of these rules should have a deterrent effect in the E.U. as they can be as high as 30 million euros or 6\% of a company annual turnover (see Article 71). \\

These regulations directly involve various applications of image analysis as they might fall into the category of high-risks systems such as medical imaging \cite{banerjee2021readingrace, Durn2021WhoIA, MUEHLEMATTER2021e195, DBLP:journals/corr/abs-1712-09923} or even facial recognition algorithm \cite{10.1145/3375627.3375820, DBLP:journals/corr/abs-2007-10075, https://doi.org/10.48550/arxiv.2208.11099} as they might relate to people daily life.

\section{Bias detection}\label{sec:BiasDetection}

Different methods were proposed to detect such (undesired) algorithmic biases in machine learning. After presenting the main Fairness metrics, we will give an overview of the recent methods that can be used to detect unknown (or non referenced) biases.

\subsection{Fairness metrics}
A large variety of fairness metrics have been introduced to quantify algorithmic biases, as presented in \cite{hardt2016equality,oneto2020fairness,DelBarrio2020ReviewFairnessML,PessachEtShmueli_ACM_2023} and references therein. They quantify different levels of relationships between a given sensitive variable $A$ and outputs of the algorithm. Yet, as fairness is a polysemous word, there exist multiple metrics, each one focusing on a particular definition of bias and, unfortunately, all of them are not necessarily compatible with each other, as recently discussed in \cite{CastelnovoEtAlNature2022} or \cite{chouldechova2020snapshot}. Therefore, it is essential for someone evaluating the bias of a model to understand what the fairness metrics really capture. They conform to different definitions of biases given in the previous section and can be decomposed as follows.

\paragraph{Statistical parity} 
One of the most standard measures of algorithmic bias is the so-called \emph{Statistical Parity}. Fairness in the sense of Statistical Parity is then reached when the model's decisions are not influenced by the sensitive variable value -- \textit{i.e. $\mathcal{L}(\hat{Y}| A) = \mathcal{L}(\hat{Y})$}. For a binary decision, it is often quantified using the Disparate Impact (DI) metric. Introduced in the US legislation in 1971\footnote{https://www.govinfo.gov/content/pkg/CFR-2017-title29-vol4/xml/CFR-2017-title29-vol4-part1607.xml} it measures how the outcome of the algorithm $\hat{Y}=f_{\theta}(X)$ depends on $A$. \\
It is computed for a binary decision as
$$ DI(f) = 
\frac{\mathbb{P}(f_{\theta}(X)=1 | A=0)}{\mathbb{P}(f_{\theta}(X)=1 | A=1)} \,,$$

where $A=0$ represents the group which may be discriminated (also called \textit{minority group}) with respect to the algorithm.
Thus, the smaller the value, the stronger the discrimination against the \textit{minority group}; while a $DI(f) = 1$ score means that Statistical Parity is reached. A threshold $\tau_0=0.8$ is commonly used to judge whether the discrimination level of an algorithm is acceptable \cite{FeldmanSIGKDD2015,ZafarICWWW17,gordaliza2019obtaining}.

\paragraph{Equal performance metrics family} 
Taking into account the input observations $X$ or the prediction errors can be more proper in various applications than imposing the same decisions for all. To address this, the notions of \emph{Equal performance}, \emph{Status-quo preserving}, or \emph{Error parity} measure whether a model is equally accurate for individuals in the sensitive and nonsensitive groups. As discussed in \cite{CastelnovoEtAlNature2022}, it is often measured by using three common metrics: \emph{Equal sensitivity} or \emph{Equal opportunity} \cite{hardt2016equality}, \emph{Equal sensitivity and specificity} or \emph{Equalized odds}, and \emph{Equal positive predictive value} or \emph{Predictive parity} \cite{ChouldechovaBD2017}. In the case of a binary decision, common metrics usually compute the difference between True Positive Rate and/or False Positive Rate for majority and minority groups. Therefore, Fairness in the sense of \textit{Equal performance} is reached when this difference is zero. Specifically:
An {\it Equal opportunity} metric is given by 
$$ |\mathbb{P}(\hat{Y}=1 | A=0, Y=1) - \mathbb{P}(\hat{Y}=1 | A=1, Y=1) |\,,$$
while an \textit{Equality of odds} metric is provided by 
$$|\mathbb{P}(\hat{Y}=1 | A=0, Y=0) - \mathbb{P}(\hat{Y}=1 | A=1, Y=0) |.$$
Finally, \textit{Predictive parity} refers to equal accuracy (or error) in the two groups. \vskip .1in
Previous notions can be written using the notion of {\bf calibration} in fairness. 
When the algorithm's decision is based on a score $s(X)$, as in~\cite{PleissEtAlNIPS2017}, a Calibration metric is defined as 
$$| \mathbb{P}(Y=1 | A=0, s(X)) - \mathbb{P}(Y=1 | A=1, s(X)) |.$$ Calibration measures the proportion of individuals that experience a situation compared to the proportion of individuals forecast to experience this outcome. It is a measure of efficiency of the algorithm and of the validity of its outcome. Yet studying the difference between the groups enable to point out difference of behaviors that would let the user to trust the outcome of an algorithm less for one group than another.  This definition extends in this sense previous notions to the multi-valued settings as pointed in \cite{barocas2017fairness}. Calibration is similar to the definition of fairness using quantile developed in \cite{yang2019fair}.

Note that previous definitions can easily be extended to the case where the variables are not binary but discrete.

\paragraph{Advanced metrics} 
First, for algorithms with continuous values, previous metrics can be understood as quantification of the variability of a mean characteristics of the algorithm, with respect to the sensitive value. So natural metrics as in \cite{benesse2022fairness} or \cite{ghosh2022biased} are given by 
$$ {\rm Var} E[f_{\theta}(X)|A] \quad \mathrm{ or } \quad {\rm Var} E[\ell(f_{\theta}(X),Y)|A]$$
Note that as pointed in \cite{benesse2022fairness}, these two metrics are unnormalized {Sobol indices}. Hence, sensitivity analysis metrics can also be used to measure bias of algorithmic decisions.  As a natural extension, sensivity analysis tools provide new ways to describe the dependency relationships between a well chosen function of the algorithm, focusing on particular features of the algorithm. They are well adapted to studying bias in image analysis.\vskip .1in

Previous measures focus on computing a measure of dependency. Yet, many authors used different ways to compute covariance-like operators, directly as in \cite{ZafarICWWW17}, or based on information theory \cite{kamishima2011fairness}, or using more advanced notions of covariance based on embedding, possibly with kernels. We refer for instance to \cite{DelBarrio2020ReviewFairnessML} for a review. Each method chooses a measure of dependency and computes a fairness measure of either the outcome of the algorithmic model or its residuals (or any appropriate transformation) with the sensitive parameter. \\
Other measures of fairness do not focus on the the mean behavior of the algorithm but other properties that may be the quantiles or the whole distribution. Hence, fairness measures compare the distance between the conditional distribution for two different values of the sensitive attribute $a \neq a'$ of either the decisions $$ d(\mathcal{L}(f_{\theta}(X)|A=a),\mathcal{L}(f_{\theta}(X)|A=a'))$$ or their loss $$d(\mathcal{L}(\ell(f_{\theta}(X),Y)|A=a),\mathcal{L}(\ell(f_{\theta}(X),Y)|A=a')).$$ Different distances between probability distributions can be used. We refer for instance to \cite{RisserEtAlJMIV2022} and references therein, where Monge-Kantorovich distance (aka Wasserstein distance) is used. Embedding of distributions using kernels can also be used as pointed out in \cite{3495724.3497012}, together with well adapted notions of dependency in this setting.

\subsection{Unknown bias detection}

We now present different methods to detect unknown bias, or more precisely, algorithmic bias with respect to sensitive variables that have to be estimated.
In essence, there are two veins in the bias detection literature: testing for the presence of a suspected bias in the model, and the discovery of sources of bias without supervision. For the former, the emphasis will be put on the structure of the trained model, whereas notions of statistical and counterfactual fairness -- with the help of generative models -- and explainability techniques will be the center point for the latter.
Although fairly new and not yet popularized in the fairness literature, the topic of bias detection is of particular interest for image-based applications, as discussed in Section \ref{ssec:causes_of_bias} and \ref{ssec:unknown_bias_in_image_analysis}. The combinatorial problem itself of identifying groups of samples without any domain knowledge or prior about what constitutes an informative representation for a specific use-case is ill-posed. This is why the techniques presented below leverage semantic information in some way or another to identify potentially discriminated groups.

In ~\cite{serna2021insidebias}, Serna et al. proposed to study the values of the activations in CNNs for a facial gender recognition task, and discover that when the models have learned a biased representation, the activations in its filters are not as high when dealing with samples from the discriminated groups. In ~\cite{serna2021ifbid}, the same group of researchers extended this work by training different groups of NNs, and then used other models to predict the presence of bias from their weights without looking at their inferences, proving that bias is encoded in the model's weights. Other works also interestingly investigate the predictor's hidden activations to detect sub-groups \cite{creager2020, sohoni2020, matsuura2019, Ahmed2021SystematicGW}.

In~\cite{denton2019detecting}, Denton et al. used generated counterfactual examples to discover and assess unknown biases. By supposing that a generative model was available, they generated counterfactual examples given a set of interpretable attributes, and tested the performance of a trained classifier. A significant drop of the classifier performance was then considered as a good indicator that a specific attribute used to generate the counterfactual example was highly influential, and could therefore reveal an unintended bias. In much the same manner, Li et al. ~\cite{li2021discover} proposed to discover unknown biased factors in a classifier by generating \textit{factor traversals} with generative models and a special hyperplane optimization. In this method, the classifier and the generative models have their weights fixed, \textit{i.e.} they are already trained, so only the hyperplane is optimized thanks to the model outputs. Thus, images traversals are generated with more and more relevance with respect to orthogonal dimensions and largest variations on the classification scores. We can also highlight the work of Paul et al. \cite{paul2021} which expand the scope of fairness from focusing solely on demographic factors to more general factors by using generative models that discover them.

By exploiting the widely known GradCAM method~\cite{GradCAM}, Tong and Kagal~\cite{tong2020investigating} recover the image classification model properties when making a decision. Their intuition is that the results could expose biases learned by the model.
For example, in a dataset where most of doctors are males, GradCAM exposed the fact that the model’s predictions focused mainly on the facial features of the person, while clothes and accessories are highlighted for female doctors. The same conclusions were drawn for basketball players, where GradCAM explained that the predictions were mainly based on the players' faces and not on basketball‐related features. Although the predictions were often accurate, explanations made indeed clear that they were based on players’ faces because the training dataset contained a lot of female volleyball players, and facial features help a lot to predict a person gender. Finally, Schaaf et al.~\cite{schaaf2021towards} combined attribution methods with ground truth masks to help detecting biases.

It is important note that the models were trained with labels in all above-mentioned methods, but biases might still be learned when using self-supervised training schemes. In~\cite{sirotkin2022study}, Sirotkin et al. looked for the presence of bias in representations learned using state-of-the-art self-supervised learning (SSL) procedures. In particular, they pre-trained models on ImageNet~\cite{deng2009imagenet} using a variety of SSL techniques and showed that there was a correlation between the type of model and the number of incorporated biases. 
Thus, they demonstrated that biases can be learned even without supervision. In a Meta-Learning fashion, \cite{Bao2022LearningTS} also propose to learn how to split a dataset such that predictors learned on a training split which cannot generalize on the test split.

\section{Obtaining fairness}\label{sec:ObtainingFairness}

Obtaining fairness of an algorithm  has been studied for a large number of applications, most of them dealing with societal problems where bias induces a potential harm for the populations, leading to potential discrimination.  Hence, mitigating the bias is focuses on obtaining algorithms which perform similarly for all groups in a population.  Although similar in some cases to the notions of fairness that are typically used in social applications -- e.g. captioning~\cite{mohler2018penalized} or predictive policy~\cite{castets2019encadrement} --, Fairness can have slightly different goals in industrial applications,.

\begin{itemize}
    \item Firstly, it is critical to obtain robust algorithms that generalize to the test domain with a certified level of performance, and that do not depend on specific working conditions or types of sensors to work as intended. The property which is sought is the robustness of the algorithm. 
    \item Secondly, the second goal is to learn representations  independent of non-informative variables that can correlate with actual predictive information and play the role of confounding variables. The relationship between Fairness and these representations constitutes an open challenge. In many cases, representations are affected by spurious correlations between subjects and backgrounds (Waterbirds, Benchmarking Attribution Methods), or gender and occupation (Athletes and health professionals, political person) that influence too much the selection of the features, and hence, the algorithmic decision. One way of studying it, is through disentangled representations~\cite{LocatelloNeurips2019}: by isolating each factor of variation into a specific dimension of the latent space, it is possible to ensure the independence with respect to sensitive variables. 
\end{itemize}

Once a source of undesirable bias has been identified, a mitigation scheme should be implemented to avoid unreliable model behaviors in certain regions of the input space. For example, it has been shown that when generating explanations on discriminated groups, the standard post-hoc explainability methods score significantly lower than when applied to samples belonging to non-discriminated groups~\cite{balagopalan2022road,dai2022fairness}. This means that fairness can be a requisite to ensure that all the properties verified by our models on majority groups are still valid for minority groups. This is particularly pertinent in industrial and safety-critical applications, where some properties can be required for the system's certification. In this case, new notions of fairness can be derived from these criteria, leading to new definitions of statistical equality implying that the properties are satisfied for all sub samples of the data.

Bias correction techniques can be divided into two categories depending on whether they are intended to be applied to problems in which the bias is already known or not. When this is the case, state-of-the-art methods mostly work by erasing the information related to the sensitive variable present in the latent space, or re-sampling/re-weighting the training dataset, or generating samples via generative models. Through the former, the latent space is split into predictive and sensitive information, and only the first part is used to learn how to predict. In contrast, by working with the training samples, the latter attempts to give more importance to under-represented groups during the training phase.

All these methods require access to the group's labels at train time, condition which might not be met on certain use-cases. It is also interesting to note that existing group labels may alternatively not be informative of actual harmful biases. Different methods were then proposed to treat these more complex cases. They can be based on either proposing potential confounding variables~\cite{seo2022unsupervised}, or on approaches from the field of Distributionally Robust Optimization (DRO)~\cite{duchi2021learning}. In particular, this latter family of methods has been a focal point for the emerging field of sub-population shift, or group shift, whereby the training distribution can be subdivided into multiple groups (oftentimes without labels) and the test distribution becomes the one of the group on which the model performs the worst.

When group labels are available at training time, the problem is well-posed and the algorithmic bias that the studied models have learned can be erased. For instance, in~\cite{zhang2018mitigating-adversarial}, Zhang et al. proposed to employ an adversarial network to modify the latent space of the classifier to optimize a given fairness metric. In a similar way, Kim et al.~\cite{kim2019learning} used an adversary to minimize the mutual information between the latent space and the sensitive variable. Grari et al.~\cite{grari2019fairness} also adversarially optimized the Hirschfeld‐Gebelein‐Rényi (HGR) maximal correlation coefficient. 
Penalty terms were also used in \cite{RisserEtAlJMIV2022,ZafarICWWW17} to ensure a good balance between model accuracy and fairness properties with respect to a sensitive variable.
In a somewhat different path, other techniques piggyback on the fact that disentangled representations can be more fair than standard ones, as the important information for the prediction is separated from the sensitive variable~\cite{LocatelloNeurips2019}. In particular, this has been applied in~\cite{creager2019flexibly,sarhan2020fairness} to split the latent space into two groups through the use of specific losses, with the task information on one side and the sensitive variable on the other.

It is also possible to encourage models to learn fair representations by only using the data. Among the simplest methods, a reweighting factor can be added to the loss so as to emphasize the discriminated samples~\cite{kamiran2012data}. An alternative is to resample the training dataset in different manners: by undersampling the dominant groups to encourage the model to learn more general rules~\cite{sagawa2020investigation}, by oversampling the discriminated groups~\cite{buda2018systematic}, by using a technique similar to MixUp~\cite{zhang2017mixup} to interpolate between dominant and minority groups~\cite{representation-neutralization}, or by generating more samples of the discriminated groups through generative models~\cite{goel2020model-patching,ramaswamy2021fair,lee2021learning}.

When the sensitive variable is not available, previous methods can not be used. The literature dealing with unknown bias is scarce, yet some solutions can be found in the machine learning literature. As in previous sections, mitigation of unknown bias amounts to correct the data or the algorithm from features that influence the algorithm. Yet from feature detection to bias mitigation, there is a gap that requires some additional knowledge that allows us to decide whether a particular direction corresponds to a bias that has to be avoided or not.

When humans are in the loop, or if the data can be described using logical variables, bias mitigation can be handled by using orthogonality constraints that prevent dependencies as in~\cite{jeon2022conservative}. When some causal information is available such as a causal graph for instance, Variational Auto-Encoders can be trained to infer some proxy for the sensitive variable as in ~\cite{grari2021fairness}. The information required is that some part of the variable are independent from the sensitive variable while the other part may be highly correlated.

Then, when the groups are unknown but the training data are known to not be completely unbiased, there are still different approaches to help improve the worst-case performance.
Namely, it is possible to propose group labels without supervision through clustering, and then apply a reweighting scheme~\cite{seo2022unsupervised}.

In all previous settings, the bias implies that the algorithm generalizes poorly to new datasets. In particular, this is the case when variations of the sensitive variable produce changes in the data distributions. Hence, algorithms that are able to achieve a good level of performance for different testing distributions are, by nature, able to handle this type of bias. Distribution Robustness of the algorithm can induce fairness in this sense. The DRO framework corresponds to solving the minimax problem $$\min_{\theta \in \Theta} \max_{Q \in \mathcal{Q}} \mathbb{E}_Q \left[\ell(Y,  f_{\theta}(X)\right]$$
where $\mathcal{Q}$ is a set of distributions close to the empirical training distribution $\mathbb{P}_n$ that we call the \textit{uncertainty set}. The choice of this $\mathcal{Q}$ establishes how the training distribution can be perturbed, with most techniques choosing all the distributions such that a f-divergence~\cite{duchi2021learning,zhai2021boosted} or a Wasserstein distance~\cite{sinha2017certifying} is smaller than a certain threshold, or by modeling it with a generative network~\cite{michel2021modeling}.  If some causal information is available  and if, so called interventions 
on the sensitive variable, can be modeled as distributional shifts on the distributions,  hence distributional robust models with respect to such shifts will be  protected from the bias induced by this variable.
Therefore, distributional robustness extends stability requirements to achieve fairness by controlling the output of the algorithm in the worst distributional case around the observed empirical distribution.

\section{A use-case for EuroSAT}\label{sec:UseCase}

\begin{figure}[h]
    \centering
    \includegraphics[width=0.99\linewidth]{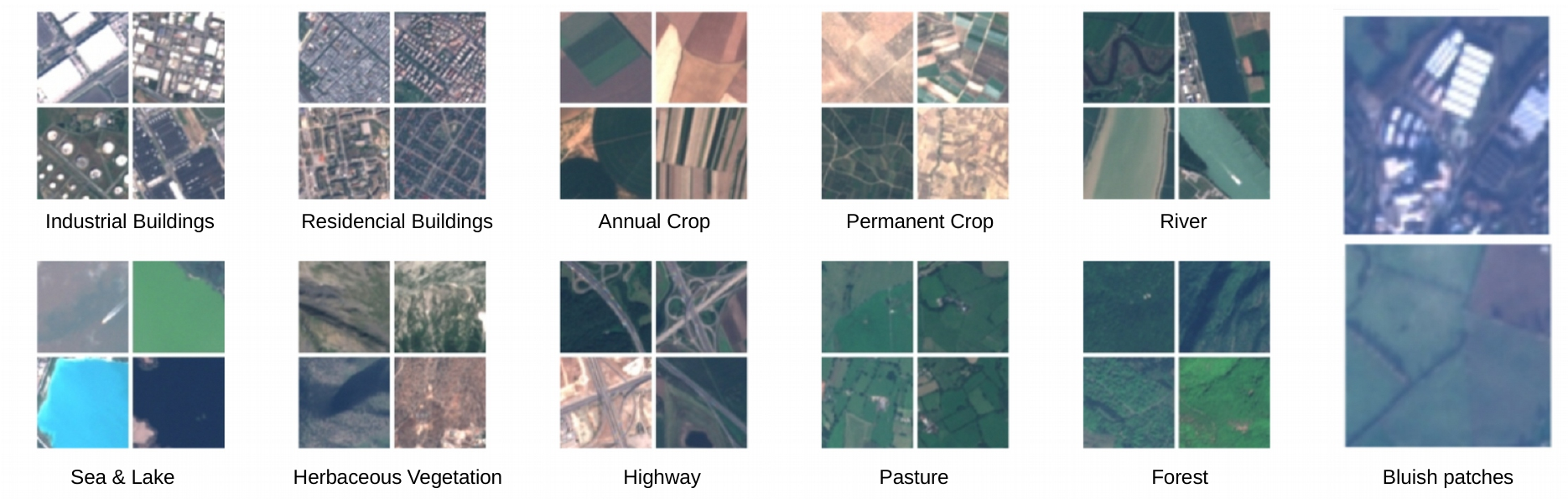} 
    \caption{
        \textbf{(Left)} Images out of the 10 classes of the EuroSAT dataset. Illustration taken from \cite{helber2019eurosat}
        \textbf{(Right)} Images of the EuroSAT dataset for which the reconstruction in the RGB color space produced the blue-veil effect.
    }
    \label{fig:BVEinEuroSAT}.
\end{figure}

We now illustrate the notions and concepts of algorithmic bias that can be encountered in industrial applications on the RGB version of the EuroSAT dataset\footnote{\url{https://madm.dfki.de/downloads}} \cite{helber2019eurosat}. It contains 27000 remote sensing images of $64 \times 64$ pixels with a ground sampling distance of 10 meters. The RGB channels were also reconstructed based on the original 13-band Sentinel-2 satellite images. Each image has a label indicating the kind of land it covers, as shown in Fig.~\ref{fig:BVEinEuroSAT}-(left).

\subsection{The blue veil effect in the EuroSAT dataset}

The blue-veil effect is caused by uncommon atmospheric conditions when acquiring Sentinel-2 images on the original 13 spectral bands, and becomes particularly noticeable when they are transformed into the visible spectrum. In essence, the picture acquires a blueish hue, as depicted in Fig.~\ref{fig:BVEinEuroSAT}-(right), that can trick classification models into thinking that it contains a mass of water. About 3\% of the dataset is corrupted by what we call the blue-veil effect. Importantly, this blue-veil effect will provide us below a typical example of algorithmic bias with an unknown sensitive variable in high-dimensional data. As we will see, blue-veil images indeed tend to be harder to classify than other images.

\subsection{Detecting sensitive variables without additional metadata}

We explain hereafter how to detect the blue-veil effect as a potential source of bias. The first step is to estimate the importance of each observation of the training dataset for a pre-trained model. This information will then be employed by clustering algorithms on specific image representations in order to automatically find the discriminated group.

We opted for the technique of \cite{picard-influence}, where first order approximations of NN influence functions were used to determine the importance of each training observation in the trained  model. With this objective in mind, a simple and generic 4-layer CNN was trained until convergence. Then, we observed that among the 25 most influential images, 7 of them were blue-veiled images, although such images only represent 3\% of the whole dataset. This suggests that training the classifier on blue-veil images is a complex task, or at least that the image features used to classify these images seem to be different than those of the other images.

\begin{figure}[h]
    \centering
    \includegraphics[width=0.99\textwidth]{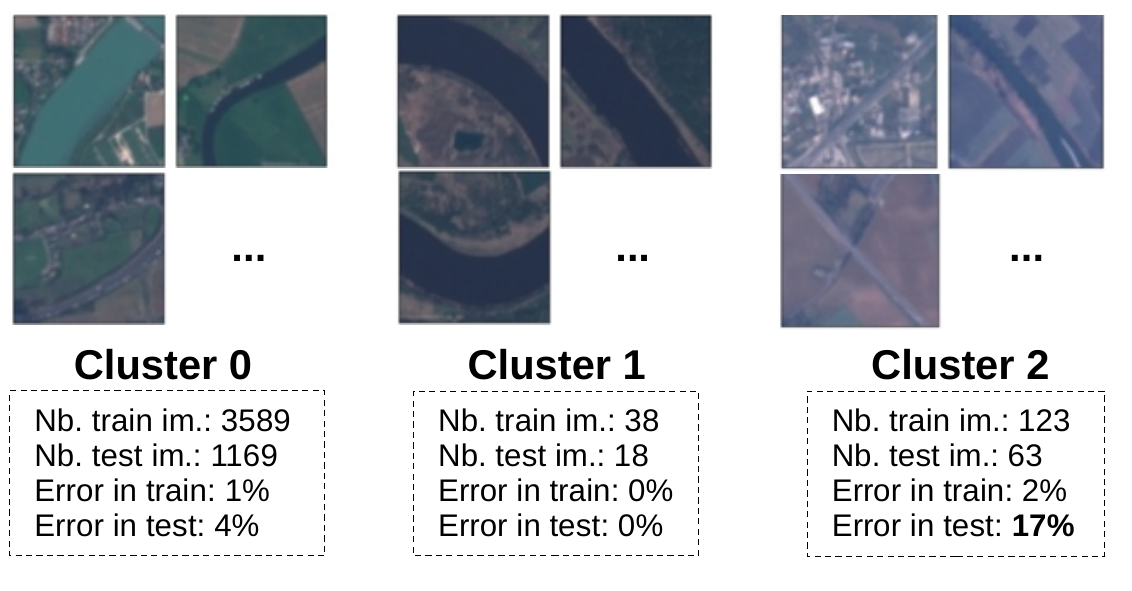}
    \caption{Detection of potentially discriminated groups and confirmation of blue-veiled images using a clustering and group-wise performance evaluation methodology. The generalization properties of a ResNet18 classifier in cluster 2, \textit{i.e.} for blue-veil images, are particularly lower than in the two other clusters.}
    \label{fig:eurosat-detection}
\end{figure}

Since the blue-veil pattern has been semi-automatically identified on several images, we can change the image representation so that a clustering algorithm will straightforwardly find and isolate all the images with this pattern in the training and test sets. For the blue-veiled images, we simply  transform the RGB (Red, Green, Blue) color space into an HSV (Hue, Saturation, Value) space, where the blue-veil images can be characterized by dominant blue colors in the \textit{Value} channel and reasonably luminous colors in the \textit{Hue} channel. By using a spectral clustering algorithm on HSV images,  we distinguish three image clusters, as shown Fig.~\ref{fig:eurosat-detection}. The first cluster contains normal-looking images, the second one mostly has large and dark rivers, and a last one represents the blue-veiled images we are looking for.

\subsection{Measuring the effect of the sensitive variable}

\begin{figure}[t]
    \centering
    \includegraphics[width=0.99\textwidth]{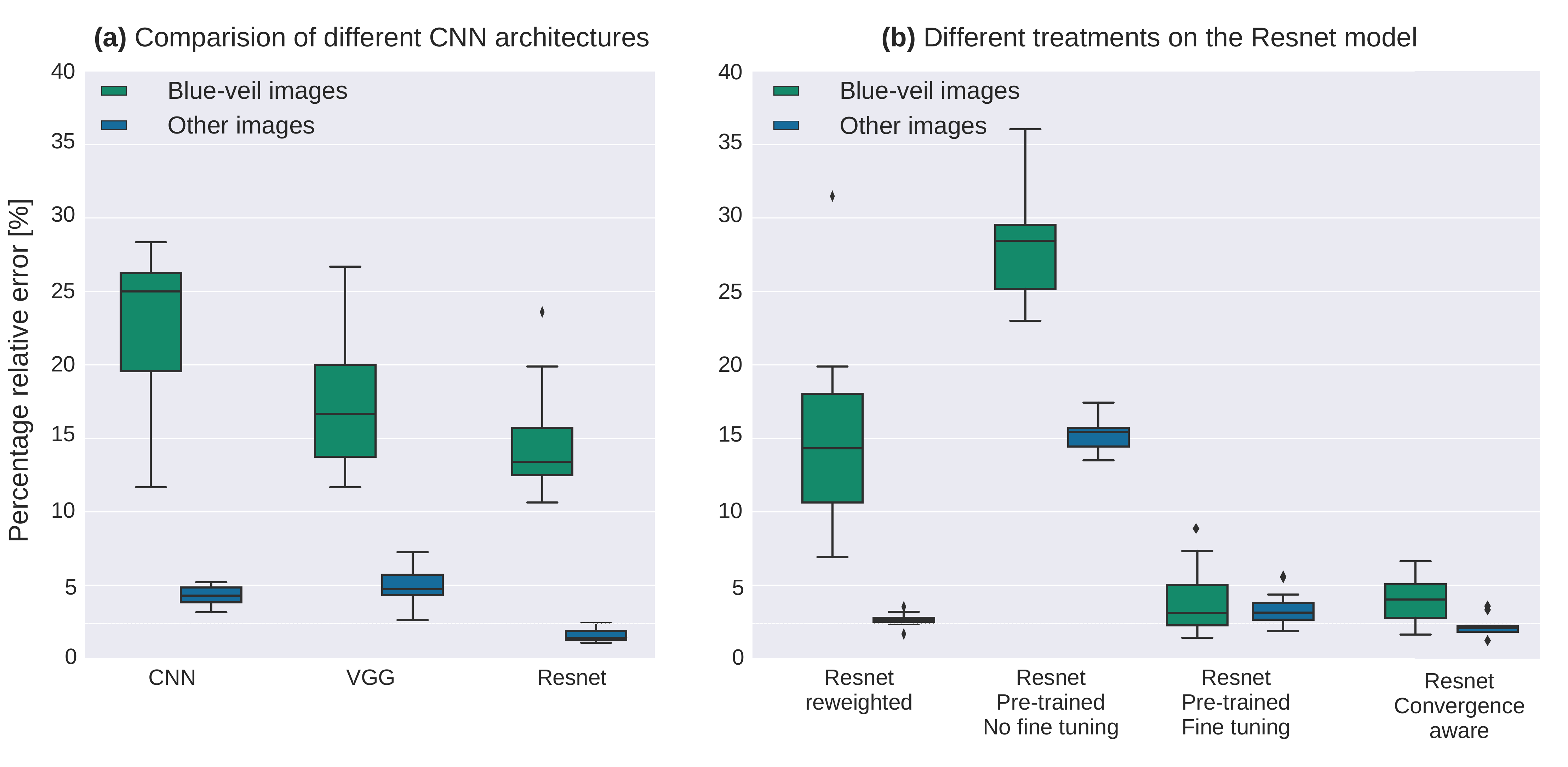}
    \caption{Average error obtained on the test set of the EuroSat dataset using different models and training strategies. For each strategy, the two boxplots distinguish the average errors on blue-veiled images (green boxplots) and other images (blue boxplots):   \textbf{(a)} Baseline results obtained on different neural-network architectures; \textbf{(b)} Effect of different treatments on the average accuracy of the Resnet architecture. 
    }
    \label{fig:merged_results}
\end{figure}

Let's check different models' performances for blue-veiled images and other images. A simple 4-layer CNN, a VGG-16 model, and a ResNet18 model were compared after being trained for 50 epochs. A total of 10 runs per configuration were used to measure the models' and learning algorithm's stability. We also focused on the binary classification between Rivers and Highways, which correspond to the two worst performing classes in the 10 class setting. This additionally forced us to train the classifiers with a more limited amount of blue-veiled images, making the problem close to what we can encounter in many industrial applications. The training and test sets indeed contained 3750 and 1250 images, respectively, where only 123 and 63 images were blueish.
As shown in Figures~\ref{fig:eurosat-detection} and \ref{fig:merged_results}-(a), we can clearly observe that the error rate is considerably higher on blue-veiled images than on other images, thus demonstrating that an undesirable algorithmic bias was learned in the sense of the equality of errors.

\subsection{Bias mitigation}

Different strategies to mitigate the undesired bias on the classifiers accuracy were also compared. All techniques tested below are based on the Resnet18 architecture, as it is the one which performed best on blue-veil images (see Fig.~\ref{fig:merged_results}-(a)). Note that we trained the models for 50 epochs by default, just like in the previous subsection. The initial parameters of the neural-networks were also randomly drawn, except in two cases that will be mentioned.

\begin{figure}[t]
    \centering
    \includegraphics[width=0.99\textwidth]{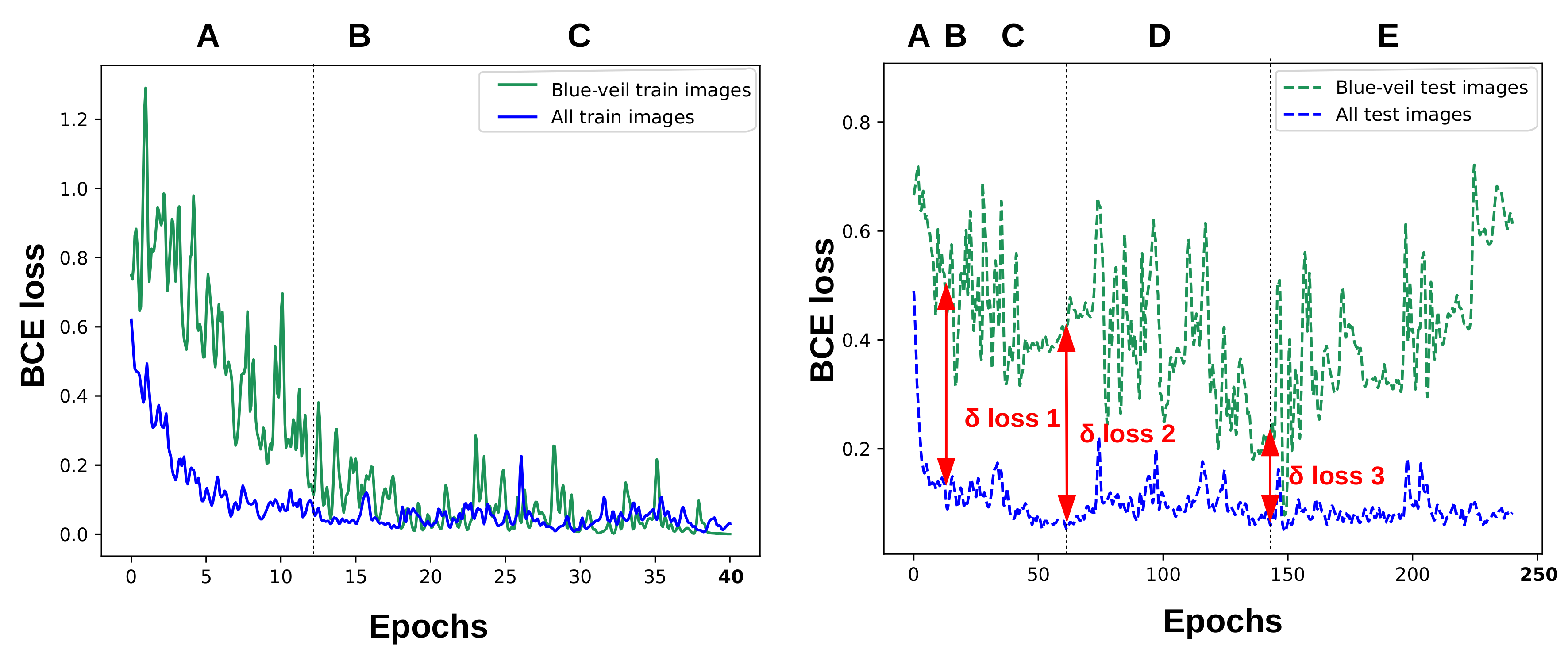}
    \caption{Detailed convergence of the BCE-loss on all data and on blue-veil images only. Results obtained on the training set \textbf{(left)} and on the test set \textbf{(right)} are represented. Note that the convergence curves obtained on the training set are only represented for the first 40 epochs, and those obtained on the test set are represented on 250 epochs. Five phases {\bf A} to {\bf E}  are distinguished to discuss the convergence behavior.}
    \label{fig:detailedConvergence}
\end{figure}

We first tested the re-weighting scheme proposed in~\cite{kamiran2012data}. The ResNet18 architecture was trained using a weighted loss, where the weights were chosen so that the disparate impact was equal to 1 (\textit{reweighted} strategy). 
We also loaded the pre-trained ResNet18 architecture of \textit{Torchvision}\footnote{\url{https://pytorch.org/vision/stable/index.html}} and  trained its last layer on our EuroSAT data to use the generic transformed image representation of this pre-trained network. It is important to mention that a very large and generic \textit{ImageNet} database was used for pre-training  (\textit{Pre-trained, No fine tuning} strategy). We alternatively fine-tuned all layers of this pre-trained network to simultaneously optimize the transformed image representation and the prediction based on this representation, \textit{i.e.} the parts 1 and 2 of the neural-network in Fig.~\ref{fig:generalNNarchitecture} (\textit{Pre-trained, Fine tuning} strategy). It is important to note that we only trained for 5 epochs instead of 50 when fine-tuning the pre-trained neural-networks in order to avoid overfitting. 
Finally, we randomly drew the initial state of the neural network and trained all layers, but thoroughly distinguished the convergence for all images and for the group of blue-veiled images only. In this case, we stopped training the ResNet18 parameters when an over-fitting phenomenon started being  observed in the blue-veiled images (\textit{Convergence aware} strategy). Results are shown in Fig.~\ref{fig:merged_results}-(b). A typical detailed convergence of the \textit{Convergence aware} strategy is also shown in Fig.~\ref{fig:detailedConvergence}.

Finally, we can discuss the results. We can first notice that the re-weighting technique of \cite{kamiran2012data} had little effect on the results. It was indeed designed to correct bias that manifests in the form of disparate impact, so it did not reduce the error rate gap between groups. The debiasing method must then be specifically chosen to target the bias through the metric with which it was measured.
Using the pre-trained network of \textit{Torchvision} had a disastrous effect when only optimizing the last neural-network layer, but was particularly efficient when using fine-tuning, \textit{i.e.} when simultaneously optimizing the transformation of the data representation and the final decision rules. Using a relevant initial state, when available, and using fine tuning then appears as a very good strategy here. It is often denoted by transfer learning in the machine learning literature.
Interestingly, similar results were obtained with a random initial state, when stopping the training procedure at an iteration where the trained neural-network had good generalization properties on the blue-veiled images. Understanding this result requires to look closely at the convergence curves, as illustrated Fig.~\ref{fig:detailedConvergence} on a typical run:

In Fig.~\ref{fig:detailedConvergence}, we compare the convergence in the whole train and test sets, as well as the blue-veil images only. We can then distinguish five phases in the convergence process. All curves start decreasing in phase {\it A}. It can only be remarked that the loss on the blue-veil images slightly increases during the 4 first epochs before decreasing, as in the average trend. We believe that this is due to a minor confounding effect. 
In phase  {\it B}, \textit{i.e.} between epochs 12 and 18, the training algorithm has converged when observed on all training images but not yet on blue-veil images. This is due to the fact that the blue-veil images only represent a small fraction of the training set. 
Note that if only measuring the convergence on the whole training set, it would be tempting to stop the training process at the beginning  of phase {\it B}, which would obviously lead to a different treatment of the blue-veiled images and other images (see $\delta$ loss 1 in Fig.~\ref{fig:detailedConvergence}).
More interestingly for us, the convergence curves are stable on the training set in phase {\it C}, but it regularly decreases on the test set. At the end of phase {\it C}, the convergence curve is stable on the whole test set, and it is common practice to stop the training process there (early stopping principle). However, it is important to remark that the generalization properties of the trained neural-network are still much poorer for blue-veiled images than other images (see $\delta$ loss 2 in Fig.~\ref{fig:detailedConvergence}). This actually explains in Fig.~\ref{fig:merged_results}-(a) the different accuracies observed for the blue-veiled images with respect to the other images. We indeed stopped the convergence after 50 epochs there.
Although noisy, the convergence curve obtained on blue-veil test images slowly decreases in phase {\it D} until reaching an optimal value at epoch 145 (see $\delta$ loss 3 in Fig.~\ref{fig:detailedConvergence}). Finally, the training algorithm starts over-fitting the blue-veiled images in phase {\it E}, so the training process should be stopped at its very beginning. It is then essential to point out that obtaining reasonably good generalization properties on blue-veiled images required about 3 times more epochs than what is made using what's commonly considered to be the good practices, and about 12 times more epochs than what would be made using a naive approach.

\section{Conclusion}

We want to put the emphasis on the fact that undesired biases may come from a variety of unexpected reasons. While in a societal applications regulations can help to focus on specific variables, it is unfortunately most of the time an arduous task which requires awareness in the data scientists community. We proposed here to go over some techniques to characterize, detect and mitigate such undesired biases. However, we also showcased the difficulty of the task when moving forward blindly. Having a strong field expertise helps to prevent some of these hidden biases as it allows to draw the attention on specific behaviors. When such an expertise is unfortunately unavailable, practitioners should even be more cautious. Some of the works presented here did not need any prior to detect and mitigate bias but oftentimes it does not allow to exhibit clearly the source of the bias, and hence, to remove the underlying cause. Use-cases leveraging AI to its full potential are growing at an exponential pace while the advances on bias characterization unfortunately move much more slowly. Thus, we tend to believe that an in-depth work is necessary before an AI-based solution is deployed, especially to preserve the end-users' trust. In that sense, we should see the different regulations arising as an opportunity to gain knowledge on data, deep learning, and optimization instead of a brake on innovation.

\section*{Acknowledgments}
    This work was conducted as part of the DEEL project\footnote{www.deel.ai}.
    Funding was provided by ANR-3IA Artificial and Natural
    Intelligence Toulouse Institute (ANR-19-PI3A-0004).


\end{document}